\documentclass[10pt,twocolumn,letterpaper]{article}

\usepackage{iccv}
\usepackage{times}
\usepackage{epsfig}
\usepackage{graphicx}
\usepackage{amsmath}
\usepackage{amssymb}

\usepackage{booktabs}
\usepackage{adjustbox}
\usepackage{subcaption}

\usepackage[pagebackref=true,breaklinks=true,letterpaper=true,colorlinks,bookmarks=false]{hyperref}

\usepackage[capitalize]{cleveref}
\crefname{section}{Sec.}{Secs.}
\Crefname{section}{Section}{Sections}
\Crefname{table}{Table}{Tables}
\crefname{table}{Tab.}{Tabs.}

\iccvfinalcopy 

\def\iccvPaperID{6987} 

\ificcvfinal\fi

\begin{document}

\title{Motion Informed Object Detection of Small Insects in Time-lapse Camera Recordings}

\author{Kim Bjerge, Carsten Eie Frigaard and Henrik Karstoft\\
Department of Electrical and Computer Engineering, Aarhus University\\
Finlandsgade 22, 8200 Aarhus N, Denmark\\
{\tt\small \{kbe, cef, hka\}@ece.au.dk}
}

\maketitle
\ificcvfinal\thispagestyle{empty}\fi

\begin{abstract}
Insects as pollinators play a crucial role in ecosystem management and world food production. 
However, insect populations are declining, calling for efficient methods of insect monitoring. 
Existing methods analyze video or time-lapse images of insects in nature, 
but the analysis is challenging since insects are small objects in complex and dynamic scenes of natural vegetation.
In this work, we provide a dataset of primary honeybees visiting three different plant species during two months of the summer period. 
The dataset consists of 107,387 annotated time-lapse images from multiple cameras, including 9,423 annotated insects.
We present a method pipeline for detecting insects in time-lapse RGB images. 
The pipeline consists of a two-step process. 
Firstly, the time-lapse RGB images are preprocessed to enhance insects in the images. 
This Motion-Informed-Enhancement technique uses motion and colors to enhance insects in images. 
Secondly, the enhanced images are subsequently fed into a Convolutional Neural network (CNN) object detector. 
The method improves the deep learning object detectors You Only Look Once (YOLO) and Faster Region-based CNN (Faster R-CNN). 
Using Motion-Informed-Enhancement, the YOLO-detector improves the average micro F1-score from 0.49 to 0.71, 
and the Faster R-CNN-detector improves the average micro F1-score from 0.32 to 0.56 on the dataset.
Our dataset and proposed method provide a step forward to automate the time-lapse camera monitoring of flying insects. 
The dataset is published on: \url{https://vision.eng.au.dk/mie/}
\end{abstract}

\section{Introduction}
\label{sec:intro}

More than half of all described species on Earth are insects, and they are the most abundant group of animals and live in almost every habitat. 
There are multiple reports of evidence for a decline in abundance, diversity, and biomass of insects in the world~\cite{Wagner2020, Didham2020, Hallmann2017, Ceballos2017}. 
Changes in the abundance of insects could have cascading effects on the food web.
Bees, hoverflies, wasps, beetles, butterflies, and moths are important pollinators and prey for birds, frogs, and bats. 
Some of the most damaging pest species in agriculture and forestry are moths~\cite{Klapwijk2013, Fox2013} and insects are known to be major factors in the world’s agricultural economy. 
Therefore, it is crucial to monitor insects in the context of global change in climate and habitats.  

Automated insect camera traps and data-analyzing algorithms based on computer vision and deep learning are valuable tools for monitoring 
and understanding insect trends and their underlying drivers~\cite{Barlow2020, Hoye2021}.
It is challenging to automate insect detection since insects move fast, and their environmental interactions, such as pollination events, are ephemeral. 
Insects also have small sizes~\cite{Hoye2021, Xia2018}, and may be occluded by flowers or leaves, making it hard to separate the objects of interest from the natural vegetation.

A particularly exciting prospect enabled by computer vision is automated, non-invasive monitoring of insects and other small organisms in their natural environment.  
Here, image processing with deep learning models of insects can be applied either in real-time~\cite{Gilpin2017} 
or batched since time-lapse images can be stored and processed after collection~\cite{Preti2021, Eliopoulos2018, Gerovichev2021, Bjerge2020, Geissmann2022}. 

Convolutional Neural Networks (CNN) are extensively used for object detection~\cite{Indolia2018, Shrestha2019, Liu2020, Zhao2019} in many contexts, 
including insect detection and species identification.
CNN for object detection predicts bounding boxes around objects within the image, their class labels, and confidence scores. 
You Only Look Once (YOLO)~\cite{Redmon2016, Bochkovskiy2020} is a one-stage object detector and has been popular 
in many applications and applied for the detection of insects~\cite{Bjerge2021}.
Two-stage detectors such as Faster Region-based Convolutional Neural Network (Faster R-CNN)~\cite{Ren2015} 
are also very common and have been adapted for small object detection~\cite{Cao2019}.

Annotated datasets are essential for data-driven insect detectors. 
Data should include images of the insects for detection and images of the typical backgrounds where such insects may be found. 
Suppose an object detector is trained on one dataset. 
In that case, it will not necessarily have the same performance on time-lapse recordings from a new monitoring site. 
One false detection in a time-lapse image sequence of natural vegetation will cause multiple false detections in the subsequent stationary images~\cite{Bjerge2020}.

We hypothesize that Motion-Information-Enhancement in insect detection in time-lapse recordings will improve the detection in the wildlife environment. 
In short, we summarize our contributions as follows:

\begin{itemize}
	\item Provide a dataset with annotated insects (primary honeybees) and a comprehensive test dataset with time-lapse annotated recordings from different monitoring sites.
	\item Propose a new pipeline method to improve insect detection in the wild, built on Motion-Informed-Enhancement, YOLOv5, and Faster R-CNN with ResNet50 as the backbone.  
\end{itemize}

\section{Related Work}
\label{sec:related}

\subsection{Detection of small objects}

Small object detection in low-resolution remote sensing images presents numerous challenges~\cite{Nguyen2020}. 
Targets are relatively small compared to the field of view, do not present distinct features, and are often grouped and lost in cluttered backgrounds.

Liu~\etal~\cite{Liu2021} compares the performances of several leading deep learning methods for small object detection.
They discuss the challenges and techniques in improving the detection of small objects. 
Techniques include fusing feature maps from shallow layers and deep layers to obtain necessary spatial and semantic information. 
Another approach is multi-scale architecture consisting of separate branches for small, medium, 
and large-scale objects generating anchors of different scales such as Darknet53~\cite{Bochkovskiy2020}.
Usually, small objects require high resolution and are difficult to recognize, 
here spatial and temporal contextual information plays a critical role in small object detection~\cite{Liu2021, Leng2021}.

A review of recent advances in small object detection based on deep learning is provided by Tong~\etal~\cite{Tong2020}.
They provide a comprehensive survey of the existing small object detection methods based on deep learning.
The review covers topics such as multiscale feature learning~\cite{Hu2018}, pyramid networks~\cite{Deng2022}, data augmentation, training strategy, and context-based detection.
Important needs for the future are proposed: emerging small object detection datasets and benchmarks, small object detection methods, and framework for small object detection tasks.

\subsection{Detection in single image}

Detection of small objects in the spatial dimension of images is investigated in several domains such as remote sensing~\cite{Ren2018} with single-shot or time-lapse images.
For small object detection tasks, the detection is very difficult since these small objects could be tightly grouped and interfere with background information. 

Du~\etal~\cite{Du2019} proposes an extended network architecture based on YOLOv3~\cite{Redmon2018} for small-sized object detection - on a complex background.
They added multi-scale convolution kernels with different receptive fields into YOLOv3 to improve extracting the semantic features 
of the objects using an Inception-like architecture inspired by GoogleNet~\cite{Szegedy2015}.

Huang~\etal~\cite{Huang2022} proposes a small object detection method based on YOLOv4~\cite{Bochkovskiy2020} for chip surface defect inspection.
They extend the backbone of YOLOv4 architecture with an enhanced receptive field by adding 
an additional fusion output ($104 \times 104$) from the Cross Stage Partial Layer (CSP2) with a similarly extended neck.

These works focus on improving the architecture for detecting small objects but are only demonstrated on a general dataset not including insects and show only minor improvements.

\subsection{Detection in a sequence of images} 

With higher framerates such as video recording~\cite{Han2021}, information in the temporal dimension can be used to improve the detection and tracking of moving objects~\cite{Bergmann2019, Zu2021}.  
The detection of small moving objects is an important research area with respect to flying insects, 
surveillance of honeybee colonies, and tracking the movement of insects.
Motion-based detections consist principally of background subtraction and frame differencing. 
State-of-the-art methods aim to combine approaches of both spatial appearance and motion to improve object detection.
Here CNNs consider both motion and appearance information to extract object locations~\cite{Sommer2021, Song2021}.

LaLonde~\etal~\cite{Lalonde2018} propose ClusterNet for the detection of small cars in wide area motion imagery. 
They achieve state-of-the-art accuracy using a two-stage deep network where the second stage detects small objects using a large receptive field.
However, the inputs are consecutive adjoining frames with frame rates of 0.8 fps. 

Stojni{\'{c}}~\etal~\cite{Stojnic2021} propose how to track small moving honeybees recorded by Unmanned Aerial Vehicles (UAV) videos.
First, they perform background estimation and subtraction followed by semantic segmentation using U-net~\cite{Weng2021} then followed by thresholding of the segmented frame.
Since a labeled dataset of small moving objects did not exist, they generate synthetic videos for training by adding small blob-like objects on real-world backgrounds.
In a final test on real-world videos with manually annotated flying honeybees, they achieved a best average F1-score of 0.71 on three small video test sequences. 

Aguilar~\etal~\cite{Aguilar22} studied small object detection and tracking in satellite videos of motorbikes.
They use a track-by-detection approach to detect and track small moving targets by using CNN object detection and a Bayesian tracker.

Insect detection and tracking are proposed in~\cite{Bjerge2021} where images are recorded in real-time with a framerate of only 0.33 fps 
performing insect detection and species classification using YOLOv3 followed by a multiple object tracker using detected center points and the size of the object-bounding box.

In-field camera recording, naturally requires hardware sufficient to process and store the videos with a real-time sampling frequency. 
This poses technical difficulties when the recording period is long and the hardware must operate without external power and network connection.
In this paper, we focus on small object detection for time-lapse recordings, which require less storage space. 
We improve insect object detection using temporal images without tracking.

\section{Dataset}

We provide a new, comprehensive benchmark dataset to evaluate data-driven methods for detecting small insects in the real natural environment.

Images in the dataset were collected using four recording units, each consisting of a Raspberry Pi 3B computer 
connected to two Logitech C922 HD Pro USB web cameras \cite{Logitech2020} with a resolution of 1920x1080 pixels. 
Images from the two cameras were stored in JPG format on an external 2TB USB hard disk. 

A time-lapse program \cite{Motion2021} installed on the Raspberry Pi was used to continuously capture time-lapse images with a framerate of 30 seconds between images. 
The camera used automatic exposure to handle light variations in the wild related to direct sun, clouds, and shadows. 
Auto-focus was enabled to handle variations of the camera distance and orientation in relation to the scene with plants and insects.
The system recorded images every day from 4:30 AM to 22:30 PM, resulting in a maximum of 2,160 images per camera per day.  

During the period May 31'st to August 5'st 2022, the camera systems were in operation in four greenhouses in Flakkebjerg, Denmark. 
The camera systems monitor insects visiting three different species of plants: \emph{Trifolium prantese} (red clover), \emph{Cakile maritima} (sea rocket), and \emph{Malva sylvestris} (common mallow). 
The camera systems were moved during the recording period to ensure different flowering plants were recorded from a side or top camera view during the whole period of observation.
A small beehive was placed in each greenhouse with western honeybees (\emph{Apis mellifera}), meaning primary honeybees were expected to be monitored during insect plant visits.  

\begin{figure}
  \centering
	 \includegraphics[width=1.0\linewidth]{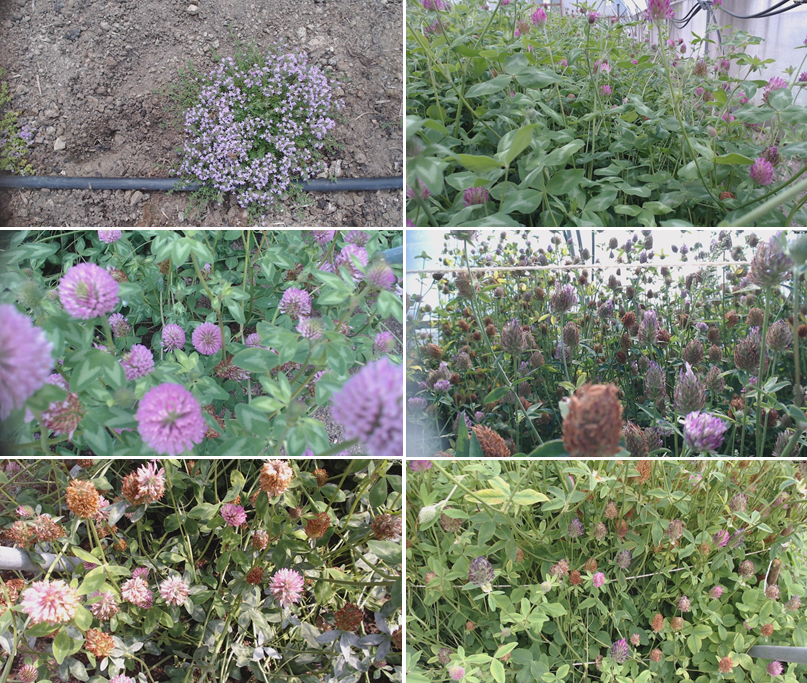}
   \caption{Example of six background images from camera systems monitoring flowing plants of sea rocket and red clover seen from a top and side view. 
	          Images were recorded by camera systems on sites shown in~\cref{tab:dataset}. (S1-1 w26, S2-1 w27, S1-1 w27, S3-0 w29, S1-0 w30, and S4-1 w29) }
   \label{fig:Backgrounds}
\end{figure}

A dataset for training and validation was created based on recordings from six different cameras with side and top views of red clover and sea rocket as shown in \cref{fig:Backgrounds}. 

Finally, a comprehensive test dataset was created by selecting seven camera sites as listed in~\cref{tab:testdata}.
The test dataset was selected to have other backgrounds and camera views than included during model training.   
The seven sites contain two weeks of recordings, monitoring common mallow, one-week monitoring sea rocket, and four weeks monitoring red clover seen from a camera top and side view.
All images were annotated using an iterative semi-automated process using human labeling and verification of model detections to find and annotate insects in more than 100,000 images. 
The goal is to evaluate the object detection models on a real dataset with another distribution than images used for training and validation.

\section{Method}
\label{sec:method}

Our proposed pipeline for detecting insects in time-lapse RGB images consists of a two-step process.
In the first step, images with Motion-Informed-Enhancement (MIE) were created.
In the second step, existing object detectors based on deep learning detectors use these enhanced images to improve the detection of small objects.  

\subsection{Motion-Informed-Enhancement}

\begin{figure}
  \centering
	 \includegraphics[width=1.0\linewidth]{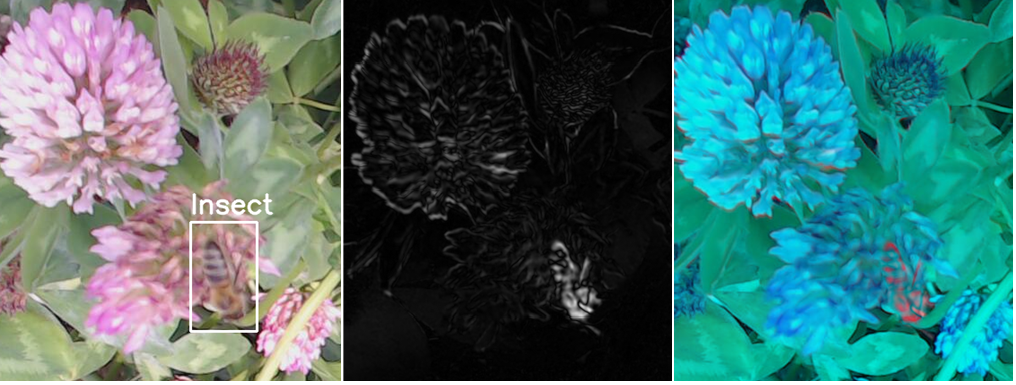}
   \caption{Left image shows the original colored image to time $k$ with a honeybee. The center image shows how the motion likelihood $3FD$ is emphasized in the image. 
	          The right image shows the motion-enhanced image $MI$ with a red color indicating information about the moving insect.}
   \label{fig:MotionImg}
\end{figure}

\begin{figure}
  \centering
	 \includegraphics[width=1.0\linewidth]{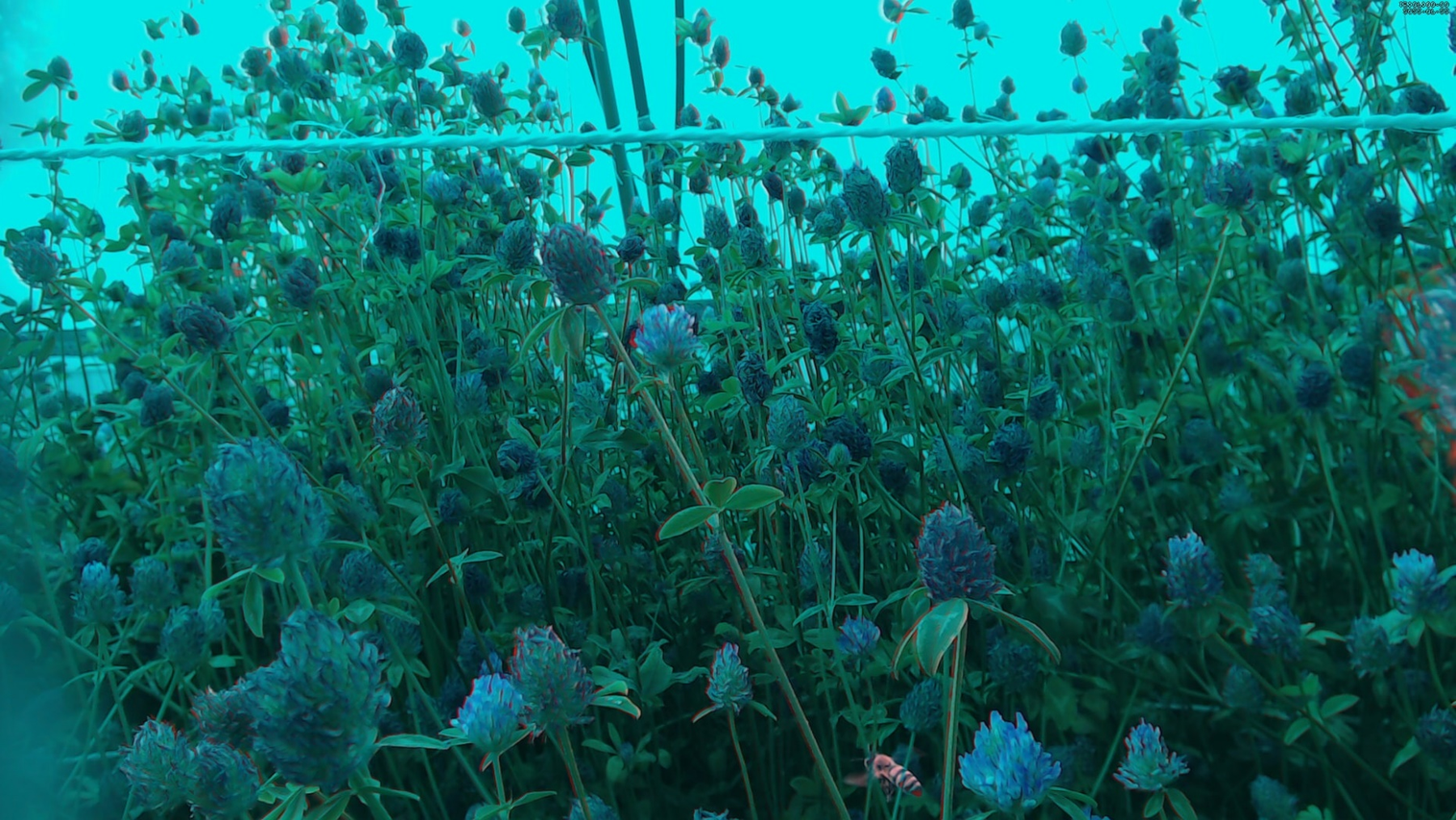}
   \caption{A full scale 1920x1080 motion-enhanced image with one honeybee.}
   \label{fig:BigMotionImg}
\end{figure}

Monitoring insects in their natural environment can be done with time-lapse cameras, where a time-lapse image is recorded at fixed time intervals of typically 30 or 60 seconds.
We hypothesize that small objects in motion will be easier to detect with deep learning detectors if images also include information from the temporal dimension in training the model.

The motion-informed detection operator proposed by Aguilar~\etal~\cite{Aguilar22} is modified for this paper to improve insect object detection using temporal images without tracking.
The detection operator estimates motion by finding the difference between consecutive frames in a time-lapse sequence. 
Our proposal modifies this method to create an enhanced image with motion information that is used for inference and training the deep learning object detector.
By using the standard RGB image format and only modifying the color content, existing object detectors can be used without modifications.
This approach can reuse popular image object detectors with CNN such as YOLO~\cite{Redmon2016} and Faster R-CNN~\cite{Ren2015}.

Three consecutive images in the time-lapse recording were used to create the enhanced image. 
The colored images were first converted to grayscale and blurred ($IGB_k$) with a Gaussian kernel of 5x5 pixels (image size: 1920x1080 pixels).  
The gray scales and blurred images were then used to create the motion likelihood $3FD_k[i,j]$, 
where $[i,j] \in [1..N] \times [1..M]$ are the pixel coordinates and $k \in \mathbb{N}$ is the time index.
This process is summarized in equations \cref{eq:delta} and \cref{eq:3FD}.
\begin{equation}
  \Delta IGB_k[i,j] = IGB_k[i,j] - IGB_{k-1}[i,j]
  \label{eq:delta}
\end{equation}
\begin{equation}
  3FD_k[i,j] = | \Delta IGB_k[i,j] | + | \Delta IGB_{k+1}[i,j] | 
  \label{eq:3FD}
\end{equation}

The original colored image, at time $k$ was then modified to create a motion-enhanced image ($MI$).
Here the enhanced blue color channel $MI_b$ was replaced by a combination of the original red ($I_r$) and blue color channels ($I_b$) shown in equation \cref{eq:blue}.
The motion likelihood $3FD$ was inserted in the enhanced red channel see equation \cref{eq:red}.
The original green channel was unchanged, copied to the enhanced green channel in equation \cref{eq:green}. 
\begin{equation}
  MI_b[i,j] = 0.5 I_b[i,j] + 0.5 I_r[i,j] 
  \label{eq:blue}
\end{equation}
\begin{equation}
  MI_r[i,j] = 3FD_k[i,j]
  \label{eq:red}
\end{equation}
\begin{equation}
  MI_g[i,j] = I_g[i,j] 
  \label{eq:green}
\end{equation}

Results of the proposed method are illustrated in \cref{fig:MotionImg,fig:BigMotionImg}. 
It shows how the motion information is created and is finally seen as a red color on the moving insect in the enhanced image.
Most of the background colors of the leaves are green and unchanged in the enhanced image. 
Colors from flowers such as pink, red, and orange are mixed in the blue channel.

\subsection{Object detection with deep learning}
Image object detection methods based on deep learning rely solely on spatial image information to extract features and detect regions of objects in the image.
In our work, Faster R-CNN with a backbone of ResNet50~\cite{He2016} and YOLOv5~\cite{GlennJocher2020} 
with a backbone of CSPDarknet53 were evaluated to detect small insects in wildlife images. 
YOLO is a one-stage object detector and Faster R-CNN is a two-stage.

One-stage object detectors predict the boundary of bounding boxes, detect whether an object is present and classify the object in the same process stage. 
One-state detectors are typically faster than two-stage detectors at the cost of lower accuracy.
However, fast execution is important when millions of images need to be processed, such as remote sensing, on a large scale.
Although remarkable results are achieved across several benchmarks, their performance decreases with small objects in complex environments such as insect monitoring.

Two-stage detectors perform region proposals before inference and classification. 
Faster R-CNN proposes a Region Proposal Network (RPN), which is a Fully Convolutional Network (FCN) that generates region proposals with various scales and aspect ratios. 
It scans the proposed regions to assess whether future inference needs to be carried out.  
The content of the proposed regions defined by a bounding box is classified in the second stage and the box coordinates are adjusted. 


In the paper~\cite{Bjerge2022} different YOLOv5 architectures are evaluated finding that YOLOv5m6 with 35.7 million parameters 
is the optimal model to detect and classify insect species in images with flowering \emph{Sedum} plants.
To improve performance and speed up training, the YOLOv5m6 and Faster R-CNN with ResNet50 are pre-trained on the COCO dataset~\cite{Lin2015}. 
For Faster R-CNN with ResNet50, a simple pipeline~\cite{Rath2022} with data augmentation was used to train the model.
The augmentation includes random vertical and horizontal image flip, image rotation, and different types of blurring. 
Images are re-sized to 1280x720 pixels for training with the two evaluated networks and transfer learning (COCO) is used to fine-tune the parameters of the CNN.

A micro- and macro-average metric was computed for the model predictions of the selected seven different physical sites in the test dataset.
The macro-average metric was computed as the average recall, precision, and F1-score for the model performance for each test site.
The micro-average aggregates the contributions from all test sites to compute metrics based on the total number of true positive, false positive, and false negative predictions.

\section{Experiment and results}
\label{sec:results}

717,311 images were recorded in the period of the experiment monitoring honeybees and other insects visiting three different plant species. 

\subsection{Train and validation}
\label{sec:train}

First, a trained model~\cite{Bjerge2022} was used to find insects in recordings from 10 different weeks and camera sites as listed in \cref{tab:dataset}.
These predictions generated a large number of images of candidate insects, which were verified.
Images with predictions were manually corrected for false positives, resulting in several images with corrected annotated insects and background images without insects. 
During quality checks, non-detected insects (false negative) were found, annotated, and added to the dataset.

\begin{table}
  \centering
  \begin{tabular}{@{}l|cccccc@{}}
    \toprule
    Cam. & Week & Days & Insects & Back. & View & Plant \\
    \midrule
     S1-1 & 26 & 2 & 1079 & 340 & Top  & Rocket \\
     S1-1 & 27 & 2 & 21   & 312 & Top  & Clover \\
     S1-0 & 29 & 7 & 395  & 143 & Top  & Rocket \\
     S1-0 & 30 & 7 & 648  & 115 & Side & Clover \\
     S2-1 & 27 & 7 & 186  & 136 & Side & Clover \\
     S3-0 & 29 & 7 & 120  & 308 & Side & Clover \\
     S4-0 & 28 & 7 & 154  & 533 & Side & Clover \\
     S4-0 & 30 & 7 & 20   & 468 & Top  & Clover \\
     S4-1 & 28 & 7 & 108  & 77  & Top  & Clover \\
     S4-1 & 29 & 7 & 83   & 93  & Top  & Clover \\
    \bottomrule
		 Total & 10 & 60 & 2,814 & 2,525 & & \\
    \bottomrule
  \end{tabular}
  \caption{Shows the camera sites and weeks from where data was selected to create a train and validation dataset.
	         System number and camera Id (Sx-0/1) identify each camera. 
					 Insects are the number of annotated insects found in the selected images.
					 Background (Back.) is the number of images without any insects where false positive detections were removed. 
					 The flowering plants are seen with a camera view from the top or side. 
					 The plant species are Sea rocket (\emph{Cakile maritima}) and red clover (\emph{Trifolium prantese}).
					 Example of background images are shown in~\cref{fig:Backgrounds}}
  \label{tab:dataset}
\end{table}

This dataset was used to create a final training dataset with an approximate split of 20\% annotations used for validation. 
The train and validation dataset were manually corrected a second time based on the motion-enhanced images and additional corrections were made. 
An additional 253 insects were found with an increase of 8\% more annotated insects compared to the first manually corrected dataset.
The datasets were created in two versions with color and motion-enhanced images.
The resulting final datasets for train and validation are listed in \cref{tab:trainval}.

\begin{table}
  \centering
  \begin{tabular}{@{}lccc@{}}
    \toprule
     Dataset & Insects & Images & Background  \\
    \midrule
     Train & 2,499 & 3,783 & 1,953 \\
     Validate & 568 & 946 & 508   \\
		\bottomrule
		 Total & 3,067 & 4,729 & 2,461 \\
    \bottomrule
  \end{tabular}
  \caption{Shows the final train and validation dataset with annotated insects and the number of images.
	         Background is the number of images without any insects.}
  \label{tab:trainval}
\end{table}

The training and validation datasets were used to train the two different object detection, Faster R-CNN with ResNet50 and YOLOv5.
The models were trained with color and motion-enhanced datasets as listed below:

\begin{itemize}
	\setlength\itemsep{-0.5em}
	\item Faster R-CNN with color images
	\item Faster R-CNN with MIE 
	\item YOLOv5 with color images
	\item YOLOv5 with MIE 
\end{itemize}

Each combination of models and dataset was trained five times.
The highest validation F1-score was used to select the best five models without over-fitting the network.
For each of the five trained models, the precision, recall, F1-score, and Average Precision (AP@.5) were calculated on the validation dataset. 
AP@.5 is calculated as the mean area under the precision-recall curve for a single class (insects) with an Intersection over Union (IoU) of 0.5. 
The average for the five trained models is listed in~\cref{tab:validresults}.

\begin{table}
  \centering
	\begin{adjustbox}{width=\columnwidth,center}
  \begin{tabular}{@{}lccccc@{}}
    \toprule
     Model & Dataset & Recall & Prec. & F1-score & AP@.5  \\
    \midrule
     FR-CNN & Color  & 0.867 & 0.889 & 0.878 & 0.890 \\
     FR-CNN & Motion & 0.889 & 0.862 & 0.875 & 0.900 \\
		 YOLOv5 & Color  & 0.888 & 0.897 & 0.892 & 0.914 \\
		 YOLOv5 & Motion & 0.919 & 0.852 & 0.884 & 0.924 \\
    \bottomrule
  \end{tabular}
	\end{adjustbox}
  \caption{Shows the average validation recall, precision, F1-score and AP@.5 for five trained Faster R-CNN and YOLOv5 models 
	         with color images and motion-enhanced images.}
  \label{tab:validresults}
\end{table}

The results show a high recall, precision, and F1-score for all models in the range of 85\% to 92\%. 
The trained models with motion-enhanced images have a recall of 1-2\% higher than with color images, but the precision is 4-5\% lower.
The trained YOLOv5 models have approximately a 1\% higher F1-score and 2\% higher AP@.5 than Faster R-CNN.
Based on the results, training with motion-enhanced images does not improve the F1-score.

\subsection{Test results and discussion}

The test dataset was created from seven different sites and weeks not included in the training and validation datasets.
A separate YOLOv5 model was trained on the training and validation dataset described in section~\cref{sec:train}. 
This model performed inference on the selected seven sites and weeks of recordings.
The results were manually evaluated, removing false predictions and searching for non-detected insects in more than 100,000 images.
In total, 5,737 insects were found and annotated in this first part of the iterative semi-automated process.
In the second part, two additional object detection models with Faster R-CNN and YOLOv5 were trained with motion-enhanced images.
These two models performed inference on the seven sites and predictions were compared with the first part of annotated images, resulting in the finding of an additional 619 insects.
The complete test dataset is listed in~\cref{tab:testdata}

\begin{table}
  \centering
	\begin{adjustbox}{width=\columnwidth,center}
  \begin{tabular}{@{}l|cccccc@{}}
    \toprule
    Cam. & Week & Insects & Images & Ratio (\%) & View & Plant \\
    \midrule
     S1-0 & 24 & 170 & 14,092 & 1.2 & Top & Rocket \\
     S1-1 & 29 & 333 & 15,120 & 2.2 & Top & Clover \\
     S2-0 & 24 & 322 & 14,066 & 2.3 & Side & Mallow \\
     S2-1 & 26 & 411 & 14,011 & 2.9 & Side & Mallow \\
     S3-0 & 28 & 2,100 & 15,120 & 13.9 & Side & Clover \\
     S4-0 & 27 & 2,319 & 15,120 & 15.3 & Side & Clover \\
     S4-1 & 30 & 701 & 15,120 & 4.6 & Top & Clover \\
    \bottomrule
  \end{tabular}
	\end{adjustbox}
  \caption{Shows the test dataset with the number of annotated insects in recordings from seven different camera sites and weeks.
	         System number and camera Id (Sx-0/1) identify each camera. 
					 The percentage ratio of annotated insects relative to the number of images recorded during each week is shown. 
	         The average ratio is 6.2\% insects based on 6,356 annotations in 102,649 time-lapse recorded images.
					 The flowering plants are seen with a camera view from the top or side. 
					 The plant species are Sea rocket (\emph{Cakile maritima}), red clover (\emph{Trifolium prantese}), and common mallow (\emph{Malva sylvestris}).}
  \label{tab:testdata}
\end{table}

The test dataset contains sites with varying numbers of insects, ranging from a ratio of 1.2\% to 15.3\% insects compared to the number of recorded images.
An average ratio of 6.2\% insects was found in 102,649 images.
Most of the annotated insects were honeybees, but a small number of hoverflies were found at camera site S1-1.
The monitoring site S1-0 of sea rocket contains other animals such as spiders, beetles, and butterflies.  
Many of the images at site S1-1 were out of focus caused of a very short camera distance to the red clover plants. 
Sites S2-0 and S2-1 monitor common mallow, which was not part of the training and validation dataset.
Site S4-0 had a longer camera distance to the red clover plants, where many honeybees were only barely visible.
In general, many insects were partly visible due to occlusion by leaves or flowers where only the head or abdomen of the honeybee could be seen. 
Additional illustrations of insect annotations and detections are included in supplementary material. 

In \cref{tab:testfrcnn} the recall, precision, and F1-score are shown are calculated as an average of the five trained Faster R-CNN models evaluated on the seven test sites.
The Faster R-CNN models were evaluated on color and motion-enhanced images.
Recall, precision, and F1-score increased for all seven test sites with Faster R-CNN models trained with motion-enhanced images.
The micro-average recall was increased by 15\% and precision by nearly 40\% indicating that our proposed method has a huge impact on detecting small insects. 
Especially verified on a test dataset with another marginal distribution than for the train and validation dataset.
The F1-score was increased by 24\% from 0.320 to 0.555. 
The most difficult test site for the models to predict was S1-0 with a low ratio of insects (1.2\%) and with animals such as spiders and beetles not present in the training dataset.

\begin{table}
  \centering
	\begin{adjustbox}{width=\columnwidth,center}
  \begin{tabular}{@{}l|cccccc@{}}
    \toprule
            & FR-CNN & Motion & FR-CNN    & Motion    & FR-CNN   & Motion    \\
     Camera & Recall & Recall & Precision & Precision & F1-score & F1-score  \\
    \midrule
     S1-0 & 0.051 & 0.262 & 0.032 & 0.758 & 0.037 & 0.385 \\
     S1-1 & 0.141 & 0.413 & 0.112 & 0.488 & 0.112 & 0.435 \\
     S2-0 & 0.305 & 0.529 & 0.250 & 0.650 & 0.274 & 0.576 \\
     S2-1 & 0.355 & 0.496 & 0.398 & 0.599 & 0.374 & 0.532 \\
     S3-0 & 0.404 & 0.487 & 0.538 & 0.840 & 0.459 & 0.612 \\
     S4-0 & 0.178 & 0.365 & 0.539 & 0.891 & 0.267 & 0.515 \\
     S4-1 & 0.496 & 0.585 & 0.262 & 0.634 & 0.337 & 0.603 \\
    \bottomrule
		 Macro & 0.276 & 0.448 & 0.305 & 0.694 & 0.266 & 0.522 \\
		 Micro & 0.300 & 0.446 & 0.344 & 0.751 & 0.320 & 0.555 \\
    \bottomrule
  \end{tabular}
	\end{adjustbox}
  \caption{Recall, precision, and F1-score on average for each camera site used in the test dataset. 
	         The average is calculated based on five trained Faster R-CNN with ResNet50 models compared 
	         with five models trained with motion-enhanced images. 
					 The macro and micro average metrics cover results from all seven camera sites and weeks.}
  \label{tab:testfrcnn}
\end{table}

\begin{figure}
  \centering
	 \includegraphics[width=0.60\linewidth]{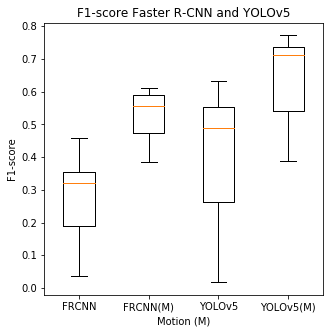}
   \caption{Box plot of F1-score for seven sites of YOLOv5 and Faster R-CNN models trained with color and motion-enhanced (M) images.
	          The horizontal orange mark indicates the micro-average F1-score based on all seven test sites.}
   \label{fig:F1scoreBoxPlot}
\end{figure}

In \cref{tab:testyolov5} the recall, precision, and F1-score are shown are calculated as an average of five trained YOLOv5 models evaluated on the seven test sites.
The YOLOv5 models were evaluated on color images and motion-enhanced images.
The micro-average recall was increased by 28.2\% and precision by only 7\%. 
However, the micro-average F1-score was increased by 22\% from 0.490 to 0.713, indicating that motion-enhanced images do increase the ability to detect insects in the test dataset.
The YOLOv5 models outperformed the Faster R-CNN trained models, achieving an increase of 16\% for the micro-average F1-score from 0.555 to 0.713.
Remark that camera sites S2-0 and S2-1 with common mallow, not included in the training, 
perform extremely well with motion-enhanced images achieving an F1-score of 0.643 and 0.618 respectively.
Camera sites S3-0, S4-0, and S4-1 achieve the best recall, precision, and F1-score.
This is probably due to the high insect ratio of 4.6\%-15.3\% and red clover plants were heavily represented in the training dataset.
\begin{table}
  \centering
	\begin{adjustbox}{width=\columnwidth,center}
  \begin{tabular}{@{}l|cccccc@{}}
    \toprule
            & YOLOv5 & Motion & YOLOv5    & Motion    & YOLOv5   & Motion    \\
     Camera & Recall & Recall & Precision & Precision & F1-score & F1-score  \\
    \midrule
     S1-0 & 0.028 & 0.284 & 0.019 & 0.693 & 0.017 & 0.389 \\
     S1-1 & 0.126 & 0.502 & 0.210 & 0.437 & 0.147 & 0.463 \\
     S2-0 & 0.288 & 0.630 & 0.619 & 0.674 & 0.376 & 0.643 \\
     S2-1 & 0.335 & 0.635 & 0.784 & 0.621 & 0.461 & 0.618 \\
     S3-0 & 0.442 & 0.694 & 0.890 & 0.879 & 0.587 & 0.772 \\
     S4-0 & 0.368 & 0.665 & 0.890 & 0.865 & 0.517 & 0.747 \\
     S4-1 & 0.486 & 0.733 & 0.917 & 0.727 & 0.634 & 0.727 \\
    \bottomrule
		 Macro & 0.296 & 0.592 & 0.619 & 0.699 & 0.392 & 0.623 \\
		 Micro & 0.377 & 0.659 & 0.718 & 0.784 & 0.490 & 0.713 \\
    \bottomrule
  \end{tabular}
	\end{adjustbox}
  \caption{Recall, precision, and F1-score on average for each camera site used in the test dataset. 
	         The average is calculated based on five trained YOLOv5 models compared 
	         with five models trained with motion-enhanced images. 
					 The macro and micro average metrics cover results from all seven camera sites and weeks.}
  \label{tab:testyolov5}
\end{table}
The box plot of the F1-scores shown in \cref{fig:F1scoreBoxPlot} indicates an increasing F1-score with motion-trained models.
It also shows a lower variation in the ability to detect insects between the seven different test sites, indicating a more robust detector.

\begin{figure}
  \centering
  \begin{subfigure}{0.9\linewidth}
	  \includegraphics[width=1.0\linewidth]{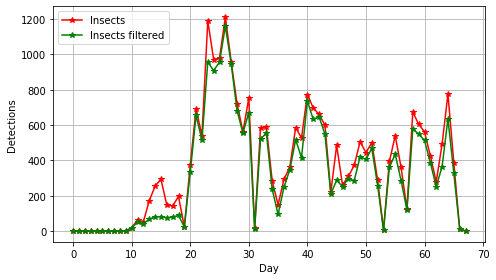}
    \caption{YOLOv5 with color images.}
    \label{fig:AbunColor}
  \end{subfigure}
  \hfill
  \begin{subfigure}{0.9\linewidth}
    \includegraphics[width=1.0\linewidth]{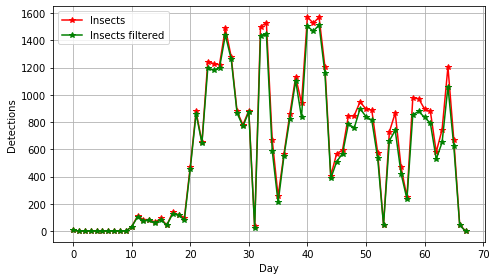}
    \caption{YOLOv5 with motion-enhanced images.}
    \label{fig:AbunMotion}
  \end{subfigure}
  \caption{The abundance of insects from the two months monitoring period of flowers and insects.
	         A two minutes filter is used to remove detections at the same spatial position in the time-lapse image sequence.
					 The red and green curve shows the non-filtered and filtered detections, respectively. 
					 The difference between the curves indicates false predictions or an insect detected at the same position within two minutes.}
  \label{fig:Abundance}
\end{figure}

\begin{figure*}
  \centering
   \fbox{\includegraphics[width=0.9\linewidth]{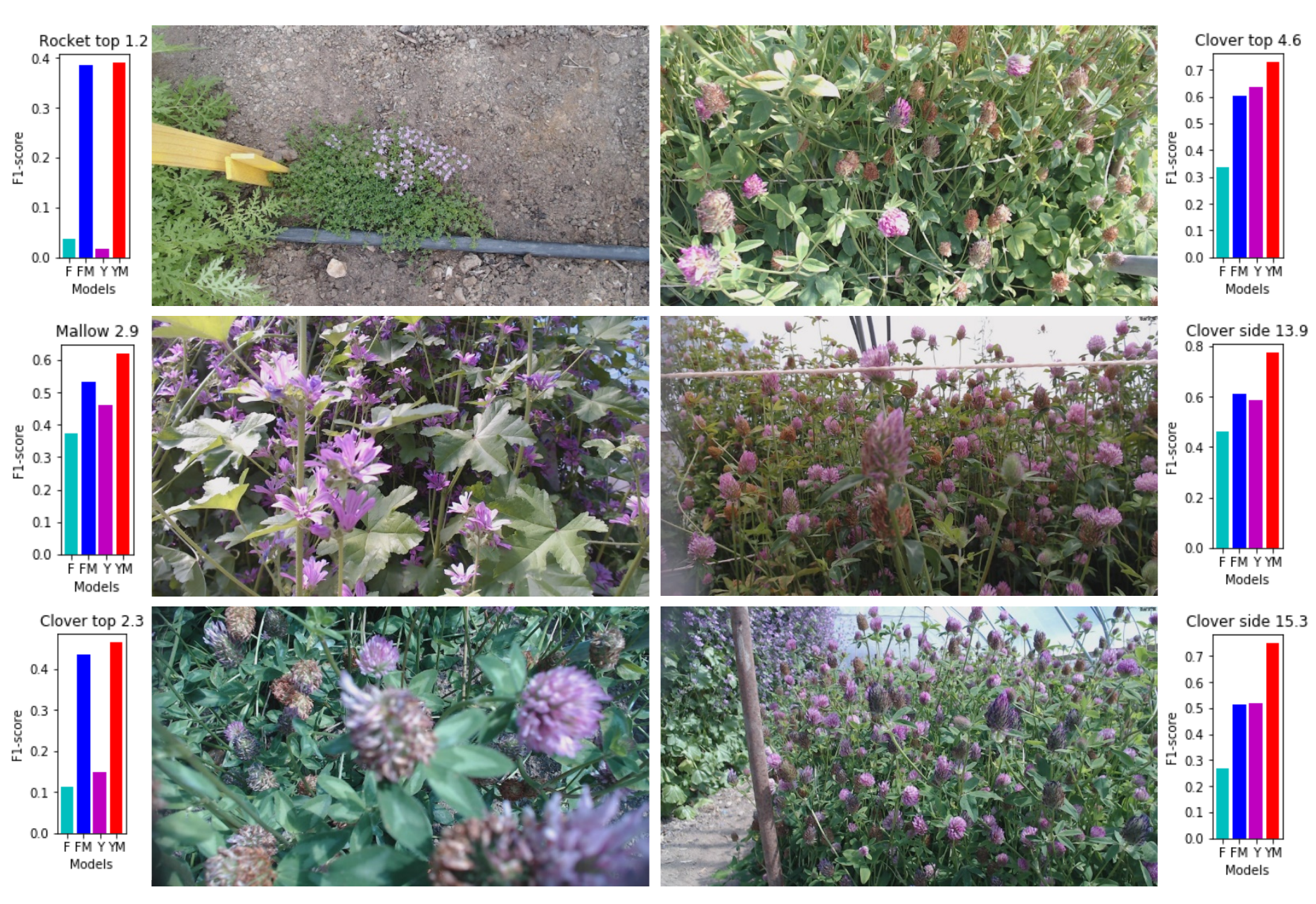}} 
   \caption{Images with micro-average F1-scores from six different sites for YOLOv5 and Faster R-CNN trained models.
	          The colors for F1-score bars of models are Faster R-CNN (F) light blue, Faster R-CNN with motion (FM) blue, YOLOv5 (Y) purple, and YOLOv5 with motion (YM) red.
						The six sites are Rocket top 1.2 (S1-0), Mallow 2.9 (S2-1), Clover top 2.3 (S1-1), Clover top 4.6 (S4-1), Clover side 13.9 (S3-0), Clover side 15.2 (S4-0).
	          }
   \label{fig:testf1score}
\end{figure*}

\cref{fig:Abundance} shows the abundance of insects detected with two YOLOv5 models trained on color and motion-enhanced images over the two months of the experiment, including images from both training, validation, and test datasets. 
False insect detections are typically found in the same spatial position of the image.
A honeybee visit within the camera view typically has a duration of fewer than 120 seconds documented in~\cite{Bjerge2021}.
A filter is therefore used to remove detections for the same spatial position within two minutes in the time-lapse image sequence. 
\cref{fig:AbunColor} shows the abundance of a YOLOv5 model trained with color images.
There are periods with a high difference in the filtered and non-filtered detections probably due to a high number of false insect detections.
\cref{fig:AbunMotion} shows the abundance of a YOLOv5 model trained with motion-enhanced images.
The model trained with motion-enhanced images shows in general a higher number of detections than the model trained with color images indicating more insects were found and detected. 
A visual overview of the results showing the micro-average F1-score for six different sites is shown in \cref{fig:testf1score}.
Here it is evident that MIE improves the ability to detect small insects with a variety of background plants, camera view, and distance.
It is also seen that with a higher ratio of insects, the overall F1-score is increased. 
Trained models with MIE are especially better at detecting insects on sites with sparse insects (Rocket top 1.2) and plants out of focus close to the camera (Clover top 2.3). 

\section{Conclusion}
\label{sec:conclusion}
This work provides a public benchmark dataset of annotated insects for time-lapse monitoring from seven different sites.
The dataset meets an important demand for future research in detecting small objects~\cite{Tong2020}. 
The test dataset includes 6,356 annotated insects in 102,649 images of complex scenes of the natural environment including three different plants of vegetation. 
A train and validation dataset is also published and verified with our newly proposed method to train the deep learning models with motion-enhanced images.
The hypothesis that Motion-Information-Enhancement will improve insect detection in the wildlife environment has been proven.
The trained CNN object detectors with YOLOv5 and Faster R-CNN demonstrate on the test datasets a micro-average F1-score of 0.71 and 0.56 respectively. 
This is a higher F1-score compared with models trained on normal color images, achieving only 0.49 with YOLOV5 and 0.32 with Faster R-CNN.
Both models trained with motion-enhanced images have a higher recall than with color images, where YOLOv5 and Faster R-CNN are increased by 28\% and 15\%, respectively. 
Our work provides a step forward to automate the monitoring of flying insects in a complex and dynamic natural environment using time-lapse cameras and deep learning.

\clearpage
\newpage
\mbox{~}

{\small
\bibliographystyle{ieee_fullname}
\bibliography{DetectionOfInsectsbib}

\begin{thebibliography}{10}\itemsep=-1pt

\bibitem{Aguilar22}
Camilo Aguilar, Mathias Ortner, and Josiane Zerubia.
\newblock {Small Object Detection and Tracking in Satellite Videos With Motion
  Informed-CNN and GM-PHD Filter}.
\newblock {\em Frontiers in Signal Processing}, 2, 2022.

\bibitem{Barlow2020}
Sarah~E. Barlow and Mark~A. O'Neill.
\newblock {Technological advances in field studies of pollinator ecology and
  the future of e-ecology}.
\newblock {\em Current Opinion in Insect Science}, 38, 2020.

\bibitem{Bergmann2019}
Philipp Bergmann, Tim Meinhardt, and Laura Leal-Taixe.
\newblock {Tracking without bells and whistles}.
\newblock In {\em Proceedings of the IEEE International Conference on Computer
  Vision}, volume 2019-October, 2019.

\bibitem{Bjerge2022}
Kim Bjerge, Jamie Alison, Mads Dyrmann, Carsten~Eie Frigaard, Hjalte M.~R.
  Mann, and Toke~Thomas Hoye.
\newblock Accurate detection and identification of insects from camera trap
  images with deep learning.
\newblock {\em bioRxiv}, 2022.

\bibitem{Bjerge2021}
Kim Bjerge, Hjalte~M.R. Mann, and Toke~T. H{\o}ye.
\newblock {Real-time insect tracking and monitoring with computer vision and
  deep learning}.
\newblock {\em Remote Sensing in Ecology and Conservation}, 2021.

\bibitem{Bjerge2020}
Kim Bjerge, Jakob~Bonde Nielsen, Martin~Videb{\ae}k Sepstrup, Flemming
  Helsing-Nielsen, and Toke~Thomas H{\o}ye.
\newblock {An automated light trap to monitor moths (Lepidoptera) using
  computer vision-based tracking and deep learning}.
\newblock {\em Sensors (Switzerland)}, 2021.

\bibitem{Bochkovskiy2020}
Alexey Bochkovskiy, Chien-Yao Wang, and Hong-Yuan~Mark Liao.
\newblock {YOLOv4: Optimal Speed and Accuracy of Object Detection}.
\newblock {\em arXiv}, 2020.

\bibitem{Cao2019}
Changqing Cao, Bo Wang, Wenrui Zhang, Xiaodong Zeng, Xu Yan, Zhejun Feng, Yutao
  Liu, and Zengyan Wu.
\newblock {An Improved Faster R-CNN for Small Object Detection}.
\newblock {\em IEEE Access}, 7, 2019.

\bibitem{Ceballos2017}
Gerardo Ceballos, Paul~R. Ehrlich, and Rodolfo Dirzo.
\newblock {Biological annihilation via the ongoing sixth mass extinction
  signaled by vertebrate population losses and declines}.
\newblock {\em Proceedings of the National Academy of Sciences of the United
  States of America}, 114(30), 2017.

\bibitem{Deng2022}
Chunfang Deng, Mengmeng Wang, Liang Liu, Yong Liu, and Yunliang Jiang.
\newblock {Extended Feature Pyramid Network for Small Object Detection}.
\newblock {\em IEEE Transactions on Multimedia}, 24, 2022.

\bibitem{Didham2020}
Raphael~K. Didham, Yves Basset, C.~Matilda Collins, Simon~R. Leather, Nick~A.
  Littlewood, Myles~H.M. Menz, J{\"{o}}rg M{\"{u}}ller, Laurence Packer,
  Manu~E. Saunders, Karsten Sch{\"{o}}nrogge, Alan~J.A. Stewart, Stephen~P.
  Yanoviak, and Christopher Hassall.
\newblock {Interpreting insect declines: seven challenges and a way forward}.
\newblock {\em Insect Conservation and Diversity}, 13(2), 2020.

\bibitem{Du2019}
Peng Du, Xiujie Qu, Tianbo Wei, Cheng Peng, Xinru Zhong, and Chen Chen.
\newblock {Research on Small Size Object Detection in Complex Background}.
\newblock In {\em Proceedings 2018 Chinese Automation Congress, CAC 2018},
  2019.

\bibitem{Eliopoulos2018}
Panagiotis Eliopoulos, Nikolaos~Alexandros Tatlas, Iraklis Rigakis, and Ilyas
  Potamitis.
\newblock {A “smart” trap device for detection of crawling insects and
  other arthropods in urban environments}.
\newblock {\em Electronics (Switzerland)}, 2018.

\bibitem{Fox2013}
R Fox, Ms Parsons, and Jw Chapman.
\newblock {The State of Britain's Larger Moths 2013}.
\newblock Technical report, Wareham, Dorset, UK, 2013.

\bibitem{Geissmann2022}
Carrillo~J {Geissmann Q, Abram PK, Wu D, Haney CH}.
\newblock {Sticky Pi is a high-frequency smart trap that enables the study of
  insect circadian activity under natural conditions.}
\newblock {\em PLoS Biol.}, 20(7), 2022.

\bibitem{Gerovichev2021}
Alexander Gerovichev, Achiad Sadeh, Vlad Winter, Avi Bar-Massada, Tamar Keasar,
  and Chen Keasar.
\newblock {High Throughput Data Acquisition and Deep Learning for Insect
  Ecoinformatics}.
\newblock {\em Frontiers in Ecology and Evolution}, 9, 2021.

\bibitem{Gilpin2017}
Amy~Marie Gilpin, Andrew~J. Denham, and David~J. Ayre.
\newblock {The use of digital video recorders in pollination biology}.
\newblock {\em Ecological Entomology}, 2017.

\bibitem{Hallmann2017}
Caspar~A. Hallmann, Martin Sorg, Eelke Jongejans, Henk Siepel, Nick Hofland,
  Heinz Schwan, Werner Stenmans, Andreas M{\"{u}}ller, Hubert Sumser, Thomas
  H{\"{o}}rren, Dave Goulson, and Hans {De Kroon}.
\newblock {More than 75 percent decline over 27 years in total flying insect
  biomass in protected areas}.
\newblock {\em PLoS ONE}, 12(10), 2017.

\bibitem{Han2021}
Yanyong Han and Yandong Han.
\newblock {A Deep Lightweight Convolutional Neural Network Method for Real-Time
  Small Object Detection in Optical Remote Sensing Images}.
\newblock {\em Sensing and Imaging}, 22(1), 2021.

\bibitem{He2016}
Kaiming He, Xiangyu Zhang, Shaoqing Ren, and Jian Sun.
\newblock {Deep Residual Learning for Image Recognition}.
\newblock In {\em Proceedings of 2016 IEEE Conference on Computer Vision and
  Pattern Recognition}, 2016.

\bibitem{Hoye2021}
Toke~T. H{\o}ye, Johanna {\"{A}}rje, Kim Bjerge, Oskar L.~P. Hansen, Alexandros
  Iosifidis, Florian Leese, Hjalte M.~R. Mann, Kristian Meissner, Claus Melvad,
  and Jenni Raitoharju.
\newblock {Deep learning and computer vision will transform entomology}.
\newblock {\em Proceedings of the National Academy of Sciences}, 2021.

\bibitem{Hu2018}
Guo~X. Hu, Zhong Yang, Lei Hu, Li Huang, and Jia~M. Han.
\newblock {Small Object Detection with Multiscale Features}.
\newblock {\em International Journal of Digital Multimedia Broadcasting}, 2018,
  2018.

\bibitem{Huang2022}
Haixin Huang, Xueduo Tang, Feng Wen, and Xin Jin.
\newblock {Small object detection method with shallow feature fusion network
  for chip surface defect detection}.
\newblock {\em Scientific Reports}, 12(1), 2022.

\bibitem{Indolia2018}
Sakshi Indolia, Anil~Kumar Goswami, S.~P. Mishra, and Pooja Asopa.
\newblock {Conceptual Understanding of Convolutional Neural Network- A Deep
  Learning Approach}.
\newblock In {\em Procedia Computer Science}, volume 132, 2018.

\bibitem{GlennJocher2020}
Glenn Jocher.
\newblock {You Only Look Once Ver. 5 (YOLOv5) on Github}, 2020.
\newblock https://github.com/ultralytics/yolov5.

\bibitem{Klapwijk2013}
Maartje~J. Klapwijk, Gy{\"{o}}orgy Cs{\'{o}}ka, Anik{\'{o}} Hirka, and Christer
  Bj{\"{o}}orkman.
\newblock {Forest insects and climate change: Long-term trends in herbivore
  damage}.
\newblock {\em Ecology and Evolution}, 3(12), 2013.

\bibitem{Lalonde2018}
Rodney Lalonde, Dong Zhang, and Mubarak Shah.
\newblock {ClusterNet: Detecting Small Objects in Large Scenes by Exploiting
  Spatio-Temporal Information}.
\newblock In {\em Proceedings of the IEEE Computer Society Conference on
  Computer Vision and Pattern Recognition}, 2018.

\bibitem{Leng2021}
Jiaxu Leng, Yihui Ren, Wen Jiang, Xiaoding Sun, and Ye Wang.
\newblock {Realize your surroundings: Exploiting context information for small
  object detection}.
\newblock {\em Neurocomputing}, 433, 2021.

\bibitem{Lin2015}
Tsung-Yi Lin, Michael Maire, Serge Belongie, Lubomir Bourdev, Ross Girshick,
  James Hays, Pietro Perona, Deva Ramanan, C.~Lawrence Zitnick, and Piotr
  Doll{\'{a}}r.
\newblock {Microsoft COCO: Common Objects in Context}.
\newblock {\em Proceedings of the IEEE Computer Society Conference on Computer
  Vision and Pattern Recognition}, 2015.

\bibitem{Liu2020}
Li Liu, Wanli Ouyang, Xiaogang Wang, Paul Fieguth, Jie Chen, Xinwang Liu, and
  Matti Pietik{\"{a}}inen.
\newblock {Deep Learning for Generic Object Detection: A Survey}.
\newblock {\em International Journal of Computer Vision}, 128(2), 2020.

\bibitem{Liu2021}
Yang Liu, Peng Sun, Nickolas Wergeles, and Yi Shang.
\newblock {A survey and performance evaluation of deep learning methods for
  small object detection}.
\newblock {\em Expert Systems with Applications}, 172, 2021.

\bibitem{Logitech2020}
Logitech.
\newblock {C922 Pro HD Stream Webcam}, 2022.

\bibitem{Motion2021}
Motion.
\newblock {Motion an open source program that monitors video from cameras.},
  2022.
\newblock https://motion-project.github.io/.

\bibitem{Nguyen2020}
Nhat~Duy Nguyen, Tien Do, Thanh~Duc Ngo, and Duy~Dinh Le.
\newblock {An Evaluation of Deep Learning Methods for Small Object Detection}.
\newblock {\em Journal of Electrical and Computer Engineering}, 2020, 2020.

\bibitem{Preti2021}
Michele Preti, Fran{\c{c}}ois Verheggen, and Sergio Angeli.
\newblock {Insect pest monitoring with camera-equipped traps: strengths and
  limitations}.
\newblock {\em Journal of Pest Science}, 94(2), 2021.

\bibitem{Rath2022}
Sovit~Ranjan Rath.
\newblock {Faster R-CNN PyTorch training pipeline}, 2022.
\newblock https://github.com/sovit-123/fasterrcnn-pytorch-training-pipeline.

\bibitem{Redmon2016}
Joseph Redmon, Santosh Divvala, Ross Girshick, and Ali Farhadi.
\newblock {You only look once: Unified, real-time object detection}.
\newblock In {\em Proceedings of the IEEE Computer Society Conference on
  Computer Vision and Pattern Recognition}, 2016.

\bibitem{Redmon2018}
Joseph Redmon and Ali Farhadi.
\newblock {YOLOv3: An incremental improvement}.
\newblock {\em arXiv}, 2018.

\bibitem{Ren2015}
Shaoqing Ren, Kaiming He, Ross Girshick, and Jian Sun.
\newblock {Faster R-CNN: Towards Real-Time Object Detection with Region
  Proposal Networks}.
\newblock In {\em Proceedings of the 28th International Conference on Neural
  Information Processing Systems}, volume~1 of {\em NIPS'15}, page 91–99,
  Cambridge, MA, USA, 2015. MIT Press.

\bibitem{Ren2018}
Yun Ren, Changren Zhu, and Shunping Xiao.
\newblock {Small object detection in optical remote sensing images via modified
  Faster R-CNN}, 2018.

\bibitem{Shrestha2019}
Ajay Shrestha and Ausif Mahmood.
\newblock {Review of deep learning algorithms and architectures}.
\newblock {\em IEEE Access}, 7, 2019.

\bibitem{Sommer2021}
Lars Sommer, Wolfgang Kruger, and Michael Teutsch.
\newblock {Appearance and Motion Based Persistent Multiple Object Tracking in
  Wide Area Motion Imagery}.
\newblock {\em Proceedings of the IEEE International Conference on Computer
  Vision}, 2021-October:3871--3881, 2021.

\bibitem{Song2021}
Shaojian Song, Yuanchao Li, Qingbao Huang, and Gang Li.
\newblock {A new real-time detection and tracking method in videos for small
  target traffic signs}.
\newblock {\em Applied Sciences (Switzerland)}, 11(7), 2021.

\bibitem{Stojnic2021}
Vladan Stojni{\'{c}}, Vladimir Risojevi{\'{c}}, Mario Mu{\v{s}}tra, Vedran
  Jovanovi{\'{c}}, Janja Filipi, Nikola Kezi{\'{c}}, and Zdenka Babi{\'{c}}.
\newblock {A method for detection of small moving objects in UAV videos}.
\newblock {\em Remote Sensing}, 13(4), 2021.

\bibitem{Szegedy2015}
Christian Szegedy, Wei Liu, Yangqing Jia, Pierre Sermanet, Scott Reed, Dragomir
  Anguelov, Dumitru Erhan, Vincent Vanhoucke, and Andrew Rabinovich.
\newblock {Going deeper with convolutions}.
\newblock {\em Proceedings of the IEEE Computer Society Conference on Computer
  Vision and Pattern Recognition}, 07-12-June-2015:1--9, 2015.

\bibitem{Tong2020}
Kang Tong, Yiquan Wu, and Fei Zhou.
\newblock {Recent advances in small object detection based on deep learning: A
  review}.
\newblock {\em Image and Vision Computing}, 97, 2020.

\bibitem{Wagner2020}
David~L. Wagner.
\newblock {Insect declines in the anthropocene}.
\newblock {\em Annual Review of Entomology}, 2020.

\bibitem{Weng2021}
Weihao Weng and Xin Zhu.
\newblock {U-Net: Convolutional Networks for Biomedical Image Segmentation}.
\newblock {\em IEEE Access}, 9:16591--16603, 2021.

\bibitem{Xia2018}
Denan Xia, Peng Chen, Bing Wang, Jun Zhang, and Chengjun Xie.
\newblock {Insect detection and classification based on an improved
  convolutional neural network}.
\newblock {\em Sensors (Switzerland)}, 2018.

\bibitem{Zhao2019}
Zhong~Qiu Zhao, Peng Zheng, Shou~Tao Xu, and Xindong Wu.
\newblock {Object Detection with Deep Learning: A Review}.
\newblock {\em IEEE Transactions on Neural Networks and Learning Systems},
  30(11), 2019.

\bibitem{Zu2021}
Shicheng Zu, Kai Yang, Xiulai Wang, Zhongzheng Yu, Yawen Hu, and Jia Long.
\newblock {UAVs-based Small Object Detection and Tracking in Various Complex
  Scenarios}.
\newblock In {\em ACM International Conference Proceeding Series}, 2021.

\end{thebibliography}
}

\end{document}